\documentclass{article}  

\usepackage[a4paper, margin=1in]{geometry}
\usepackage{subfig,xcolor,helvet,courier,amsmath,mathtools,comment, hyperref}
\usepackage[linesnumbered,ruled,vlined]{algorithm2e}

\DeclareMathOperator*{\argmax}{arg\,max}

\newcommand{\irow}[1]{\begin{smallmatrix}(#1)\end{smallmatrix} }
\newcommand{\icol}[1]{\left(\begin{smallmatrix}#1\end{smallmatrix}\right) }

\newcommand{\target}{\mathcal{M}^{target}}
\newcommand{\setsource}{\mathcal{T}_{sub}}
\newcommand{\mdpk}{\mathcal{M}^{sub}_k}

\usepackage{amsthm}
\theoremstyle{plain}
\newtheorem{thm}{Theorem}
\theoremstyle{definition}
\newtheorem{defn}[thm]{Definition}

\begin{document}

\title{Faster Reinforcement Learning Using Active Simulators}  



\author{
Vikas Jain\\
Indian Institute of Technology Kanpur\\
\texttt{vksjn18@gmail.com}
\and
Theja Tulabandhula \\
University of Illinois Chicago\\
\texttt{tt@theja.org}
}
\maketitle

\begin{abstract}
In this work, we propose several online methods to build a \emph{learning curriculum} from a given set of target-task-specific training tasks in order to speed up reinforcement learning (RL). These methods can decrease the total training time needed by an RL agent compared to training on the target task from scratch.  Unlike traditional single-task transfer learning, we consider creating a sequence from several training tasks in order to provide the most benefit in terms of reducing the  total time to train. Our methods utilize the learning trajectory of the agent on the curriculum tasks seen so far to decide which tasks to train on next. An attractive feature of our methods is that they are weakly coupled to the choice of the RL algorithm as well as the transfer learning method. Further, when there is domain information available, our methods can incorporate such knowledge to further speed up the learning. We experimentally show that these methods can be used to obtain suitable learning curricula that speed up the overall training time on discrete and continuous task domains.
\end{abstract}



\section{Introduction}

In reinforcement learning (RL), the knowledge obtained from training on a source task can be transferred to learn a target task more efficiently \cite{taylor2009transfer}. A \emph{source task} is typically similar to and sometimes less complex than the target task. A natural extension to this idea is to transfer knowledge across a sequence of source tasks and then to the target task. When such a sequence of  source tasks is not readily available, one typically has to build a few candidate tasks from scratch and  hope that the total time  spent to  train on (a subset of) these \emph{training tasks} and then the target is less than the time  needed to  learn on the target task directly.

Narvekar et al.~\cite{narvekar2016source,narvekarijcai17} present preliminary attempts in direction. They define the problem of \emph{curriculum learning} for RL as follows: Design a sequence of tasks (i.e., a curriculum) on which a learning agent learns sequentially by transferring knowledge across the stages of the curriculum,  ultimately leading to reduced learning time on the target task. 

There are two parts to any solution for curriculum learning: the first is to create a set of source tasks, and the second is to determine a curriculum from these. Though in \cite{narvekar2016source,narvekarijcai17}, the authors showed that a curriculum can speed up the learning on the target task, they stop short of addressing the process of finding a curriculum from the source tasks. Instead, they focus on creating agent-specific training tasks from the target task. In this work, we address the second key aspect of the curriculum learning problem, viz., given a set of training tasks, how can one order them in an \emph{online} fashion to form a  curriculum that speeds up the overall training for a target task. 

We propose methods for online learning of a curriculum in two scenarios: first, when there is no domain knowledge about the tasks at hand, and second, where there is additional knowledge about the tasks. In the latter case, such domain information can be encoded in various ways (for instance, in \cite{narvekar2016source}, the authors are able to design a parametric model for tasks using domain knowledge). We use the available domain information as features to extrapolate learnability across training tasks, which consequently is used to further speed up curriculum learning.
In both settings, our proposed methods are independent of the choice of RL algorithm employed by the agent as well as the choice of the transfer technique used (the latter can depend on the representation maintained by the RL agent, viz., \emph{Q}-values, policies or the model estimates).  We only assume that the RL algorithm and the transfer technique used are compatible with each other. Thus our methods can work with existing RL algorithms and transfer methods introduced in the literature~\cite{fachantidis2013transferring,lazaric2012transfer}.
In \cite{svetlik2017automatic}, the authors also attempt to create a curriculum automatically using domain information and a notion of transfer potential. While we also give methods that do not need domain information, in the case when domain information is available, we create a simple path rather than a directed acyclic graph. Further, they rely on potential based reward shaping whereas we rely on estimating learning improvements directly from the agent's experience without controlling the rewards that it sees.

\section{Related Work} \label{RELATED}

Offline curriculum learning in the context of supervised learning has been explored before. In ~\cite{bengio2009curriculum}, the authors showed that learning using a curriculum has an effect on the rate of convergence of the prediction model. In~\cite{pentina2015curriculum}, the authors proposed an approach that processes multiple supervised learning tasks in a sequence and finds the best curriculum of tasks to be learned. This is analogous to our reinforcement learning setting, although the emphasis in ~\cite{pentina2015curriculum} is on empirical risk minimization rather than training time. Active learning methods have also been used to find a curriculum of supervised learning tasks in the \emph{lifelong machine learning} setting~\cite{ruvolo2013active,pang2014smart}. However, as mentioned before, all these methods apply to the supervised learning setting and typically exploit domain knowledge. In contrast, for reinforcement learning, as we show in our experiments, domain knowledge is not necessary for online curriculum learning and can still lead to faster training of an agent on the target RL task.

Analogous to our work is that of ~\cite{zhu16}, wherein the authors define a  student-teacher setting for supervised learning. There, the teacher's task  is to provide a sequence of training data to the supervised learning algorithm  such that the latter can quickly learn the correct hypothesis.  As we show later, our methods  follow the template of  presenting a task to the RL agent for a number of steps  and then changing the task (for instance, picking a slightly more difficult but transfer-learnable task) such that the agent is steered towards the target task.  

Transfer in reinforcement learning addresses the problem of designing schemes for efficient knowledge transfer from a given source task to a given target task~\cite{barreto2016successor,taylor2011introduction,taylor2008autonomous}. However, the focus here is typically on efficient transfer (which in itself is a hard problem) and not on the selection of source tasks. Certain recent works in multi-task reinforcement learning look at the problem of task selection where the agent is presented with a set of tasks to learn from \cite{wilson2007multi}. Here, the goal is to make the agent learn policies that are simultaneously good for all the tasks. As such, this is very different from our setting where we want our agent to learn a target task successfully in a time efficient manner. 

In ~\cite{sinapov2015learning}, the authors presented an offline method to find a curriculum by computing transfer measure for each pair  of training tasks to form an inter-task transferability model. They then use this model to find a curriculum in a recursive fashion offline. Not only this is computationally intensive, such an approach is bound to fail in our online setting where we measure the success of learning a curriculum based on the total time to train (i.e., the time to learn the curriculum and the time to train on the target task). The computation of the inter-task transferability model requires a significant amount of time and domain knowledge. In that time, one could potentially train an RL agent on the target task directly from scratch. Nonetheless, when there is domain information available in the form of features, we are able to build on this approach by proposing an active learning method. This method of ours builds a model similar to that in ~\cite{sinapov2015learning} while benefiting from reduced computation time.  

In theory, designing a curriculum while designing the training tasks in a coupled manner  can further improve the total time needed for an agent to learn a target task. Unfortunately, the current state of the art is not yet amenable to a coupled approach.  
In this work, we reasonably assume a set of training tasks similar to the target task are already present and we only need to create a curriculum and learn on the target task faster than learning on the target directly from scratch. 
Finally, note that both \cite{matiisen2017teacher} and \cite{narvekarijcai17} use POMDP/MDP framework to design a curriculum, and need to specify the dynamics and rewards of the curriculum designer. In contrast, by creating estimators of transfer directly using agent's experience, we are able to design principled sequential approaches that perform well, as shown in our experiments.


\section{Preliminaries} \label{PRELIM}

\noindent\textbf{Transfer in RL}: In transfer reinforcement learning, instead of learning on a task $\mathcal{M}$ directly, an agent first learns on a suitable source task $\mathcal{M}_{sub}$. The learned knowledge (for instance, \emph{Q}-values) from $\mathcal{M}_{sub}$ is then reused to warm-start learning on the target task $\mathcal{M}$ making learning the latter task faster. Different works propose different ways of transferring the knowledge from the source task $\mathcal{M}_{sub}$ to $\mathcal{M}$ including reusing the samples~\cite{lazaric2008transfer}, the learned policy~\cite{fernandez2006probabilistic}, model of the environment~\cite{fachantidis2013transferring} and the value function~\cite{taylor2005behavior}. In our work, we demonstrate the  performance of our online curriculum learning methods using \emph{Q}-function transfer. This entails initializing \emph{Q}-function for the agent working with $\mathcal{M}$ using those obtained for $\mathcal{M}_{sub}$. In the literature, there are ways to facilitate such a transfer even if the state and action spaces of the two tasks are distinct, as long as there is a cross-task mapping between them \cite{taylor2007transfer}.\\

\noindent\textbf{Active learning}: We discuss active learning in the well known setting of ordinary least squares (with additive Gaussian noise assumption). In this problem, the objective is to find a model that minimizes the Mean Squared Error (MSE) between the true values and the predicted values. That is, we minimize over $\theta$, the objective: $ \sum_{t=1}^{n} (y_t - \theta^T \icol{x_t\\1})^2 $, where $(x_t, y_t)_1^n$ is the training data. An analytical solution to this problem (under certain conditions) is given by $\hat{\theta} = (X^TX)^{-1}X^Ty $, where $X$ is a matrix whose rows are indexed by training examples $\irow{x_t^T&1}$ ($1\leq t\leq n$) and $y$ is a column vector indexed by training example ground truth values $\irow{y_t}$.
Further, let $\theta^*$ be the true parameter that linearly relates $y_t$ and $x_t$. Then, the expected Euclidean distance between the true and the estimated parameters is given by $$ E(\|\hat{\theta} - \theta^*\|^2) = \sigma^{*2}\textrm{Tr}[(X^TX)^{-1}]$$ where $\sigma^{*2} $ is the noise variance and $\textrm{Tr}[\cdot]$ is the trace operator. 

In the active learning variant of this problem~\cite{jaakkola}, we sequentially select training examples to minimize the estimation error. Let's assume that we already have some training points (so $X$ is well defined) and let $A = (X^TX)^{-1}$. If we now choose a training example $\irow{x^T&1}$ and  append it to $X$, then $A$ matrix gets updated as: $$ \icol{X\\x^T\ 1}^T\icol{X\\x^T\ 1} = (X^TX) + \icol{x\\1}\icol{x\\1}^T = A + vv^T $$ where $v = \irow{x^T&1}^T$. Since we want to choose the training example such that the estimation error is minimized, we can write estimation error in terms of the new example:
$$ Tr[(A+vv^T)^{-1}] = Tr[A] - \frac{v^TAAv}{1+v^TAv}. $$ And observe that we only need to choose a $v$ that maximizes $\frac{v^TAAv}{1+v^TAv}$. If we assume that our data is normialized (and hence $\|v\| = 1$), then the maximizing $v$, $\hat{v}$, turn out to be the normalized eigenvector of $A$ with the largest eigenvalue. We can thus choose the next training point that is most similar to $\hat{v}$.

\section{Problem Setup} \label{PROBLEM}

Let the MDP corresponding to the target task be $\target$.  Let the  set of training tasks be  $\setsource = \{\mdpk:1\leq k\leq K\}$, where each $\mdpk$ is an MDP similar to $\target$ with possibly different state and action spaces.

The objective of curriculum learning is to design an algorithm that, when given an RL agent as an input, constructs a (sub)sequence of tasks from $\setsource$ and trains the agent on them first and then on the $\target$ to achieve a desired level of performance. And in particular, we want the total number of steps it takes for designing the curriculum, training the agent on the curriculum and then on the target $\target$, to be lesser than the steps needed to train the agent directly on $\target$ for the same level of performance. In other words, the algorithm (active simulator) actively chooses and switches the tasks that the RL agent is training on. It is assumed that the RL agent is able to transfer knowledge from previous tasks while starting a new task. The algorithm itself can be agnostic to the exact transfer technique. Let the (sub)sequence of tasks designed by the algorithm be $\mathcal{M}^*_1 \rightarrow \mathcal{M}^*_2 \rightarrow \cdots \rightarrow \mathcal{M}^*_i \rightarrow \target$ where $\mathcal{M}^*_j \in \setsource$ $1\leq j\leq i$. Further let the ordered set of curriculum tasks be denoted by $\mathcal{I}_m = (\mathcal{M}^*_1, \mathcal{M}^*_2, \cdots, \mathcal{M}^*_i)$.
We next provide the definition of curriculum in the context of our work.
\begin{defn}
 Given a target task $\target$ and a set of training tasks $\setsource = \{\mdpk:1\leq k\leq K\}$, where each $\mdpk$ is an MDP similar to $\target$ with possibly different state and action spaces, a curriculum is defined by an ordered set of tasks  $\mathcal{I}_m = (\mathcal{M}^*_1, \mathcal{M}^*_2, \cdots, \mathcal{M}^*_i)$ where $\mathcal{M}^*_j \in \setsource$ $1\leq j\leq i$, such that when an agent is trained in the order $\mathcal{M}^*_1 \rightarrow \mathcal{M}^*_2 \rightarrow \cdots \rightarrow \mathcal{M}^*_i \rightarrow \target$, the agent desired performance is attained faster than training directly on $\target$.
\end{defn}


An offline variant of the curriculum selection problem can be shown to be NP-hard via a reduction to the seriation problem~\cite{fogel2013convex}. 

We hypothesize that learning the target task after transfer from a learned curriculum is faster than learning on the target task alone or by using single stage transfer methods. Intuitively, the algorithms that address online curriculum learning have to find a learning curriculum (a sequence) from these training tasks and make the agent learn on the sequence by transferring knowledge from one task to another, with the ultimate aim of learning the target task in the lowest number of simulation steps. Because these algorithms change the task (\emph{i.e.}\ the simulation environment) seen by the agent over time, we call them \emph{active simulators} (see Algorithm \ref{AS} for a template). 
The algorithms may also initially preprocess and store information that might facilitate finding a curriculum.

The algorithms are given a total budget of $T$ steps within which they need to train the RL agent on the target task, where each step signifies one state--action--reward--state cycle. All our algorithms operate in two phases: curriculum selection and training on training tasks, and training on the target task. The first phase has a budget of $g(T)<T$ steps that can optionally be provided as an input. The algorithms also have access to an oracle $TLearn(\mathcal{L}, \mathcal{M})$, where $\mathcal{L}$ is an RL agent and $\mathcal{M}$ is an RL task.
This oracle is an abstraction to the process where the RL agent trains on a task for some number of steps. At the end of this training, the oracle returns the reward accumulated by $\mathcal{L}$ and the number of steps it trained for.

\begin{algorithm}
\SetAlgoLined
\KwIn{$\mathcal{L}$, $\setsource = \{\mdpk\}_{k=1}^{K} $, $\target$, $T$}
\KwOut{$\mathcal{L}$}
\hrule
$g(T) = 0$\\
$\mathcal{I}_m \leftarrow \phi$\\ 
$\tau \leftarrow \textrm{PreProcess}(\target$, $\setsource$, $\mathcal{L})$\\
$g(T) \leftarrow g(T) + \tau$\\
\While {$\setsource \neq \phi$}{
	$\mathcal{M}^*$, $\tau_1 \leftarrow \textrm{SelectNextTaskToLearn}(\target$, $\setsource$, $\mathcal{L})$\\
	$\mathcal{L}$, $\tau_2 \leftarrow \textrm{Tlearn}(\mathcal{L}$, $\mathcal{M}^*)$\\
	$\mathcal{I}_m \leftarrow \mathcal{I}_m \cup \mathcal{M}^*$\\ 
	$g(T) \leftarrow g(T) + \tau_1 + \tau_2$\\
	$\setsource \leftarrow \setsource \setminus \mathcal{M}^*$\\
}
$\mathcal{L} \leftarrow \textrm{TLearn}(\mathcal{L}$, $\target$, $T - g(T))$\\
\Return $\mathcal{L}$
\caption{Active Simulators\label{AS}}
\end{algorithm}

Further, the algorithms have no control over $\mathcal{L}$'s internals (say a tabular method or a function approximation based method) and treat it as a black box. One of the consequences of this is that we cannot reset the learned knowledge of the agent. To get around this, we assume that our algorithms can clone the agent at any given step to freeze the learned knowledge (more details below). We also assume that the agent $\mathcal{L}$ can first transfer previously learned knowledge (say in the form of a policy or \emph{Q}-values or the dynamics) using a transfer method and then learn a given task $\mathcal{M}$.

The obtained curriculum and an agent learned on the curriculum need to be evaluated against the case where the agent is learned only on the target task. In the case of curriculum learning, \emph{Time to Threshold} evaluation criteria \cite{taylor2009transfer} seems to be an appropriate choice for evaluating the proposed curriculum, where the time to reach a threshold performance metric is measured. The total time taken by the agent to reach a threshold performance on a target task by learning through a curriculum should ideally be less than directly learning on the target task. It might be possible that for a particular task, a threshold performance cannot be defined. In this case, \emph{Total Reward} evaluation criteria \cite{taylor2009transfer} can be used for the evaluating the proposed curriculum, where the total reward accumulated by an agent is measured. The total reward accumulated by an agent on the target task by learning through a curriculum should ideally be more than directly learning on the target task. We used both of these evaluation criteria in our experiments.

In the next section, we present several variants of \emph{Active Simulators}, where each differs in how it implements the \textrm{SelectNextTaskToLearn} method (see Algorithm \ref{AS}).
\section{Active Simulators for Online Curriculum Learning} \label{METHOD}
We now present several active simulator variants under two scenarios, viz., the domain agnostic and the domain aware settings. In the domain agnostic setting, online curriculum building is based on task specific rewards accumulated and their transformations. In the domain aware setting, curricula are obtained using features that describe each of the tasks. The active simulator methods are: (a) Reward Maximizing Greedy Selector, (b) Local Transfer Maximizing Selector, (c) Active Reward Maximizing Greedy Selector, and (d) Active Local Transfer Maximizing Selector. Among them, the first two are domain agnostic and the latter two are domain aware methods.

\subsection{Reward Maximizing Greedy Selector} \label{secRM}
    
	In this method (see Algorithm \ref{RM}), the next task in the curriculum is selected greedily based on the performance of the agent on each task in the set of remaining tasks to choose from (in Algorithm \ref{RM}, the performance is captured by total reward collected and is denoted by $\mathcal{M}_i.r$). In order to achieve this, the method needs access to multiple copies (clones) of the RL agent $\mathcal{L}$ so that it can accurately measure the transfer effect. This is because, when the agent trains on a task $\mathcal{M}_u$ and then transfers to train on another task $\mathcal{M}_v$ (for instance, to measure the transferability between the two tasks), the information state of the agent at the end of training on $\mathcal{M}_v$ is different from the information state at the beginning of training. For a third task $\mathcal{M}_w$, one cannot now compute its transferability with $\mathcal{M}_u$ unless the state of the agent after training on $\mathcal{M}_u$ is preserved. The use of clones helps the method replicate an agent's knowledge level for computing inter-task transferability reliably. Note that making a copy is a milder assumption compared to having direct access to the internal state and architecture of the RL agent.
    
\begin{algorithm}
	\SetAlgoLined
	\KwIn{$\mathcal{L}$, $\setsource = \{\mdpk\}_{k=1}^{K'}$, $\target$}
	\KwOut{Task $\mathcal{M}^*$, time taken $\tau$}
	\hrule
    $\tau \leftarrow 0$\\
	\For {each $\mathcal{M}_i \in \setsource$}{
		$\mathcal{L}_c \leftarrow $ copy($\mathcal{L}$)\\
		$\mathcal{M}_i.r$, $\tau_i \leftarrow$ \textrm{EvaluateTask}($\mathcal{L}_c$, $\mathcal{M}_i)$\\
		$\tau \leftarrow \tau + \tau_i$\\
	}
	$\mathcal{M}^* \leftarrow \argmax_{\mathcal{M}_i \in \setsource} \mathcal{M}_i.r$\\
	\Return $\mathcal{M}^*$, $\tau$
	\caption{\emph{SelectNextTaskToLearn}: Reward Maximizing Greedy Selector  \label{RM}}
\end{algorithm} 

Thus there are multiple learners: (a) the original agent $\mathcal{L}$, and (b) virtual learners $\{\mathcal{L}_c\}$ that copy agent $\mathcal{L}$'s state. A copy agent $\mathcal{L}_c$ is used to assess inter-task transferability effect starting from the current state of $\mathcal{L}$. The method then selects the task where the copy agent gets the maximum reward and adds it to the curriculum. Going back to Algorithm~\ref{AS}, the original agent $\mathcal{L}$ then learns from this new task that was added to the curriculum. This process is repeated till all the subtasks are added sequentially into the curriculum. After training on all tasks in the curriculum, the method trains the agent $\mathcal{L}$  on the target task $\target$. One key aspect of this method is that this selects the next task based on the sequence of tasks already in the curriculum at every decision epoch. Also, this procedure does not require any preprocessing step before learning on the curriculum. Hence, the $\textrm{PreProcess}$ method for this selector is a \emph{no-op}.

	In Algorithm \ref{RM}, $\tau_i$ is the number of steps used for evaluating inter-task transferability. Procedure $\textrm{EvaluateTask}(\mathcal{L}_c, \mathcal{M}_i)$ outputs the reward accumulated on task $\mathcal{M}_i$ by $\mathcal{L}_c$ in $\tau_i$ number of steps. Note that, this procedure does not alter the original RL agent $\mathcal{L}$. The length of the curriculum obtained in this method is equal to the number of training tasks $K$. 
    
\subsection{Local Transfer Maximizing Selector} \label{secIT}

In this variant (see Algorithm \ref{IT}), the effect of transfer from one task to another task is calculated for each task pair in $\setsource$ as well as for pair of tasks where the first task is in $\setsource$ and the second task is $\mathcal{M}_{target}$. Based on the $K\times(K+1)$ numbers computed during the preprocessing step (see Algorithm \ref{IT-preprocess}), the method designs a curriculum that maximizes local transfer. That is, the curriculum is obtained by greedy search starting from $\mathcal{M}_{target}$ and traversing in the reverse order. In particular, a predecessor task with which the transfer effect to $\mathcal{M}_{target}$ is the highest is first selected. Then, the process is repeated recursively and a sequence of tasks $\mathcal{M}^*_1 \rightarrow \mathcal{M}^*_2 \rightarrow \ldots \rightarrow \mathcal{M}_{target}$ is obtained. Then the method assigns this curriculum to the RL agent $\mathcal{L}$. As in the previous method, copies of the agent $\mathcal{L}$ are used to accurately capture the transfer effect.

\begin{algorithm}
	\SetAlgoLined
	\KwIn{$\mathcal{L}$, $\setsource = \{{\mathcal{M}^{sub}}_k\}_{k=1}^{K}$, $\target$}
	\KwOut{Transferability matrix $F$, time taken $\tau$}
	\hrule
    
	$F[K][K+1] \leftarrow 0$, $\tau \leftarrow 0$\\
	\For {$\mathcal{M}_i \in \setsource$}{
    	\For {$\mathcal{M}_j \in \setsource \cup \mathcal{T}_{target}$}{
        $\mathcal{L}_{c} \leftarrow $ copy($\mathcal{L}$)\\
		$F[i][j]$, $\tau_{ij} \leftarrow \textrm{TransferMeasure}(\mathcal{L}_c, \mathcal{M}_i, \mathcal{M}_j)$\\
        $\tau \leftarrow \tau + \tau_{ij}$\\
		}
	}
	\Return $F$, $\tau$
	\caption{\emph{PreProcess}: Local Transfer Maximizing Selector \label{IT-preprocess}}
\end{algorithm}

\begin{algorithm}
	\SetAlgoLined
	\KwIn{$\mathcal{L}$, $\setsource = \{{\mathcal{M}^{sub}}_k\}_{k=1}^{K'}$, $\target$, $F$}
	\KwOut{$\mathcal{M}^*$, $\tau$}
	\hrule
	$\tau \leftarrow 0$\\
	$C \leftarrow \{\target\}$\\
	$NextTask \leftarrow \mathcal{M}_{target}$\\
	\While{$\setsource \neq \phi$}{
	$\mathcal{M}_{PrevTask} \leftarrow \argmax_{\mathcal{M}_i \in \setsource} F[\mathcal{M}_i][NextTask]$\\
	$C \leftarrow C.append(\mathcal{M}_{PrevTask})$\\
	$NextTask \leftarrow \mathcal{M}_{PrevTask}$\\
	}
	$C \leftarrow \textrm{Reverse}(C)$\\
	$\mathcal{M}^* \leftarrow C.front$\\
	\Return $\mathcal{M}^*$, $\tau$
	\caption{\emph{SelectNextTaskToLearn}: Local Transfer Maximizing Selector \label{IT}}
\end{algorithm}

In Algorithm~\ref{IT-preprocess}, the quantity $F$ is the inter-task transferability matrix where each entry $F[i][j] = \textrm{TransferMeasure}(\mathcal{L}_c, \mathcal{M}_i, \mathcal{M}_j)$ stores the reward accumulated while training on task $\mathcal{M}_j$ after transferring knowledge (a learned policy or \emph{Q}-values etc.) from experience with task $\mathcal{M}_i$.
Note that the copy agent $\mathcal{L}_c$ also follows $\mathcal{L}$'s transfer method while capturing transferred knowledge. The method spends $\tau_{ij}$ steps for computing the transfer effect for each pair of tasks. The length of the curriculum obtained in this method is $K$, which is the size of the training task set $\setsource$. This method is in principle an offline method similar to \cite{sinapov2015learning} but without taking domain knowledge in consideration.


\subsection{Active Reward Maximizing Greedy Selector}

This is the first of the two domain aware methods we propose. In this setting, we are also given $n$-dimensional feature vectors $f_i = [f_{i1}, f_{i2}, \ldots, f_{in}]$ for each task $\mathcal{M}_i \in \setsource$.
 
We will adapt the Reward Maximizing Greedy Selector method described in Algorithm \ref{RM} to this information rich setting in the following way. Consider the partial list of tasks ($\mathcal{M}^*$) at a particular iteration $m$ in the while loop and denote it by $\mathcal{I}_m$. When we need to decide on the next task in the curriculum at step $m+1$, we will avoid evaluating the reward collected (see line 4 in Algorithm \ref{RM}) for every possible task under consideration. 

In other words, instead of finding the transferability measure for each sequence-task pair (the sequence being $\mathcal{I}_m$ and the task being one of the remaining tasks in $\setsource$ at step $m+1$), the method will now use the task specific features to create feature vectors for every pair of tasks and build a transferability prediction model. In particular, the method will apply the active learning technique described in Section \ref{PRELIM} to build a regression model and predict the transferability measure for sequence-task pairs. Further, it will relearn the model at each step $m$. 
	
	Let the curriculum sequence at step $m$ be $\mathcal{I}_m = \{\mathcal{M}^*_1, \allowbreak \mathcal{M}^*_2, \ldots,\allowbreak \mathcal{M}^*_{m-1}\}$. Feature vectors for the sequence as well as the tasks that are not yet in the curriculum are designed as follows. The feature vector $f^{\mathcal{I}_m}$ for sequence $\mathcal{I}_m$ is computed as: $ f^{\mathcal{I}_m}_k = \frac{1}{m-1}\sum_{i = 0}^{m-1}f^*_{ik}$, where $f^*_i$ is feature of task $\mathcal{M}^*_i$. For each remaining task in $\setsource$ (say $\mathcal{M}_j$), the sequence-task pair feature vector $f^{\mathcal{I}_mj}$ is consequently defined as:
	$ f^{\mathcal{I}_mj}_k = \frac{f_{\mathcal{I}_mk} - f_{jk}}{\max(f_{\mathcal{I}_mk}, \epsilon)} $ for $k=1,...,n$, where $\epsilon > 0$ is a small positive constant.
	
To find a new task to be added to the curriculum, a regression model is learned at each step. At step $m$, the method learns a regression model from $\mathcal{I}_m$ and the remaining tasks in $\setsource$. Then, it picks that task in $\setsource$ which has maximum value of transferability measure. The length of the curriculum obtained in this method is $\leq K$, because one may not include a particular task based on the regression predictions.

\subsection{Active Local Transfer Maximizing Selector}

 The second domain aware method extends Local Transfer Maximizing Selector (Algorithm \ref{IT}) as follows.	Instead of finding transferability measure $F[i][j]$ for every pair of tasks $\forall i,j$, it uses the features of the tasks to form pair wise feature vectors and learn a regression model $\hat{F}$ that can predict the entries of matrix $F$. Again an active learning technique can be used to build such a model using limited number of task  pairs $(i,j)$, speeding up the curriculum design in the process.
	
The feature vector $f^{ij}$ of a task pair is computed accounting for the similarity of the two tasks' features. The $k$-th element $f^{ij}_k$ of the vector $f^{ij}$ is defined as: $ f^{ij}_k = \frac{f_{ik} - f_{jk}}{\max(f_{ik}, \epsilon)}$, where $\epsilon >0$ is a small constant. The regression model $\hat{F}$ is used instead of inter-task transferability matrix $F$ and the algorithm functions very similar to Algorithm \ref{IT}. The length of the curriculum obtained in this method can also be $\leq K$. 

We conclude by summarizing the differences between the two settings. Firstly, since there is no task representation available to domain agnostic methods, we need to measure the transfer performance for all task pairs (or sequence-task pairs). On the other hand, in domain-aware scenarios, one can utilize the task features to model and predict the task pair (or sequence-task pair) transfer effects and avoid computing them for all pairs. Thus, tools such as active learning can be used (as discussed above) to reduce the number of pairs required to learn a transfer effect model, leading to potential gains in the number of steps needed while learning and training on a curriculum. 


Second, the domain independent methods proposed above cannot easily prune the training task set while designing a curriculum. It may be possible that some of the training tasks in $\setsource$ are redundant. Identifying this redundancy can help reduce the total steps spent on the curriculum before starting the training on $\target$. If task features are available, then one could potentially compute a diversity score $d(\mathcal{M}, \mathcal{I}_m)$, which measures diversity (inverse of similarity) between task $\mathcal{M}$ and the tasks in task-sequence $\mathcal{I}_m$. Using such a score, one can prune potentially redundant tasks from the curriculum. In particular, methods can be extended to do the following: they can decide to add a new task $\mathcal{M}$ into the curriculum $\mathcal{I}_m$ only if the diversity score $d(\mathcal{M}, \mathcal{I}_m)$ is greater than a threshold. This will led to curricula that are less sensitive to the number of tasks in the training task set $\setsource$.

\section{Experiments and Results} \label{RESULT}

We verified the effectiveness of our methods on three domains: (a) a maze environment \cite{asada1994vision}, (b) a grid world environment, and (c) a cart pole environment \cite{barto1983neuronlike}. The maze and grid world environment domains have discrete state and action spaces while the cart-pole domain has a continuous state space and a discrete action space. We found that the curricula selected by our approaches indeed outperform the alternative (training on the target) in all cases. In the domain aware setting, we also demonstrate that active learning methods also speed up training on the target task considerably. 

\subsection{Maze Environment}
\begin{figure}[h]
\centering
\begin{tabular}{c@{\hskip 0.4in}c}
{\includegraphics[scale=0.25]{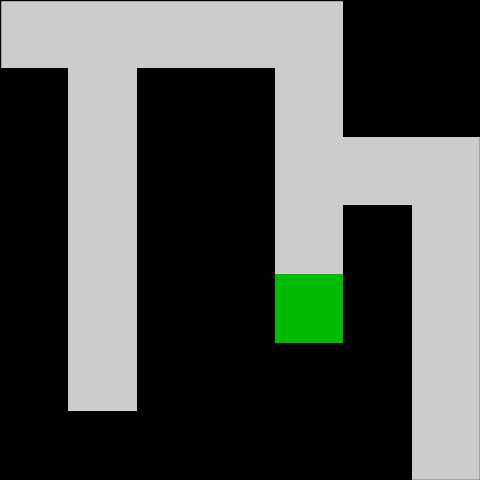}}&
{\includegraphics[scale=0.25]{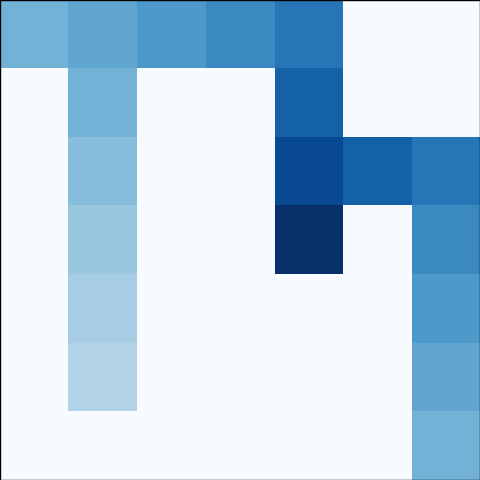}}
\end{tabular}
\caption{\small \textit{Left:} Target task in the maze environment with goal position (\emph{green}), feasible positions (\emph{gray}) and blocked positions (\emph{black}). \textit{Right:} State-Value table learned after convergence for the maze task. Darker shades represent numerically higher values.}
\label{fig:maze}
\end{figure}

In this domain (see Figure \ref{fig:maze} \textit{left}), a maze in the form of an $n\times n$ grid is provided to an agent, whose objective is to reach a specified goal state starting from a randomly chosen feasible state (gray blocks). This is an episodic task where the episode ends when the agent reaches the goal state. The agent receives a positive reward on reaching the goal state and no reward otherwise. If an agent is unable to reach the goal state within $200$ steps, the episodes ends with zero reward.
\begin{figure}[h]
\centering
\begin{tabular}{cccc}
{\includegraphics[scale=0.15]{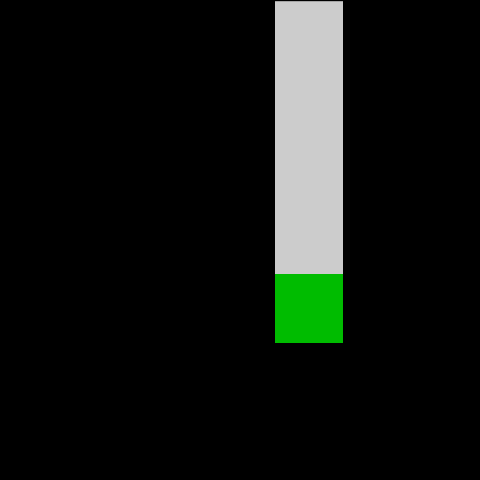}}&
{\includegraphics[scale=0.15]{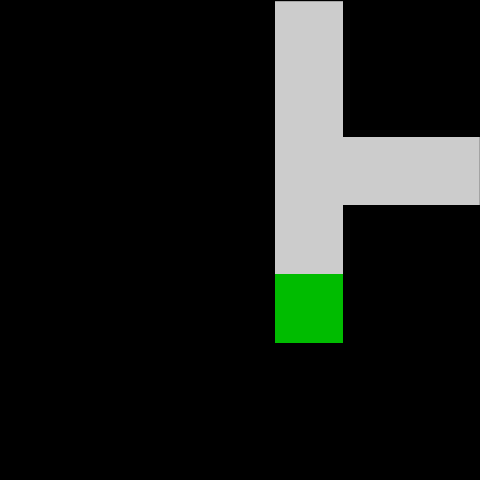}}&
{\includegraphics[scale=0.15]{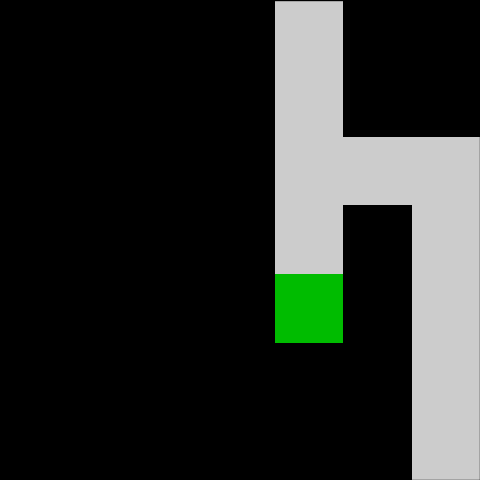}}&
{\includegraphics[scale=0.15]{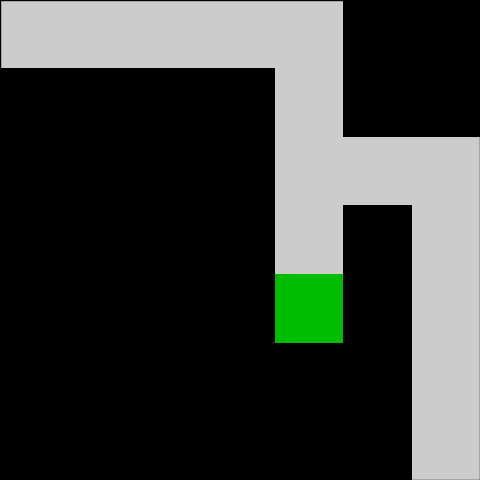}}
\end{tabular}
\caption{\small Training tasks for the maze task in Figure \ref{fig:maze}.}
\label{fig:maze-sources}
\end{figure}

In the experiment, we used the maze environment illustrated in figure \ref{fig:maze} \textit{(left)}, which represents the final target task. We created a set of source tasks $\mathcal{T}$ of size $4$ consisting of similar, but less complex tasks compared to the target task (see Figure \ref{fig:maze-sources}). The source tasks in this domain are created by reducing the state space of the task while keeping the goal position invariant. The agent employs the $Q$-learning algorithm and uses $Q$-function transfer \cite{qtransfer} as the transfer method. That is, the agent initializes itself with learned $Q$-values of a previous learned training task at the start of learning on the next task. We set the discount factor ($\gamma$) to be $0.9$ and learning rate to be $0.6$ in the $Q$-learning algorithm. 

In this domain, the target task is learned till convergence. We assume that the $Q$-values have converged if they do not change for 5 consecutive episodes. In \textit{Reward Maximizing Greedy Selector}, the number of steps for evaluating a task to be chosen as the next, is set to $200$; and the training tasks in the curriculum are learned till their convergence. In \textit{Local Transfer Maximizing Selector}, the number of steps for calculating the transfer measure between any two tasks is set to $300$ and the source tasks in the curriculum are learned till convergence.

\begin{figure}[h]
\centering
\includegraphics[scale=0.45]{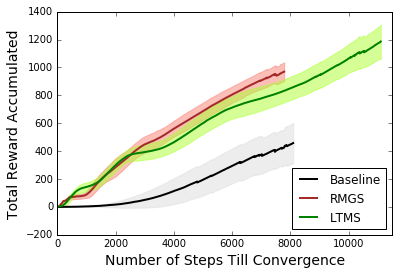}
\caption{\small Reward accumulated over the learning task versus the number of steps needed for convergence for each of the proposed methods (abbreviated in the legend) and the baseline on the maze task.}
\label{fig:maze-result}
\end{figure}

\begin{figure}[h]
\centering
{\includegraphics[scale=0.45]{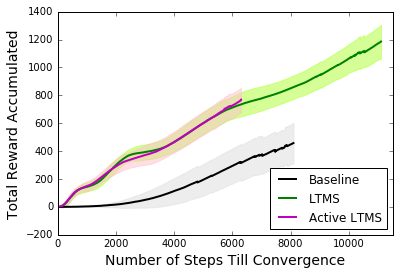}}
\caption{Comparison of Active LTMS method with the LTMS method and baseline shown in Figure \ref{fig:maze-result} on the maze task.}
\label{fig:maze-active}
\end{figure}

Figure \ref{fig:maze-result} shows the results of our active simulator algorithms and the baseline. For this domain, we evaluate our methods with the baseline based on the \textit{time to threshold} (see Section \ref{PROBLEM}) evaluation criterion where the threshold performance is defined when the agent converges on the target task. We plot the total reward accumulated by our methods as well as the baseline until the \emph{Q}-function converges on the target task (with Monte Carlo averaging over 30 runs). The state-value function learned using the curriculum is shown in Figure \ref{fig:maze} \textit{(right)}. In Figure~\ref{fig:maze-result}, it can be seen that \textit{Reward Maximizing Greedy Selector} converges in less number of steps than the baseline while \textit{Local Transfer Maximizing Selector} takes more number of steps than the baseline. To address this, we use the active learning method described in Section \ref{PRELIM} to estimate the inter-task transferability matrix using a smaller number of task pairs. For this domain, the inter-task feature $f^{ij} = [1, f^{ij}_{1}]$ for any two tasks $\mathcal{M}_i$ and $\mathcal{M}_j$ is defined to be the number of overlapping states between the two tasks divided by the total number of feasible states in the task $\mathcal{M}_i$:
$$ f^{ij}_1 = \frac{|\mathcal{S}(\mathcal{M}_i) \cap \mathcal{S}(\mathcal{M}_j)|}{|\mathcal{S}(\mathcal{M}_i)|},$$
where $\mathcal{S}(\mathcal{M})$ is the set of states of task $\mathcal{M}$. Figure \ref{fig:maze-active} shows that the number of steps needed reduce significantly using \textit{Active Local Transfer Maximizing Selector} while the performance and curricula selected remain the same.

\begin{figure}[h]
\centering
{\includegraphics[scale=0.4]{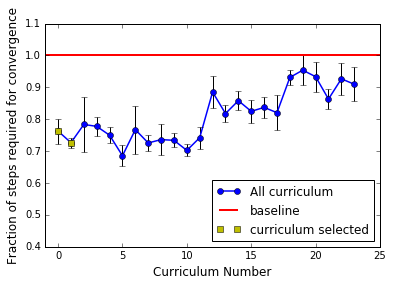}}
\caption{\small Number of steps required for convergence per curriculum (indexed from $0$ to $23$), compared with the baseline(\emph{red}) for the maze task. The curricula represent all permutations of tasks in Figure \ref{fig:maze-sources}, ordered lexicographically. \emph{Yellow} squares represent the most frequent (mode) curricula selected by our methods.}
\label{fig:maze-curr}
\end{figure}

Finally, we qualitatively analyze the curriculum proposed by the our methods. We compare all possible full-length curricula from the four source tasks (in total $24$).  Figure \ref{fig:maze-curr} shows the convergence time of each curriculum as compared to the baseline and the most frequent curriculum selected by our methods. The ordering corresponding to the most frequent curriculum is shown in order (left to right) in Figure~\ref{fig:maze-sources}.

\subsection{Grid World Environment}
\begin{figure}[h]
\centering
\begin{tabular}{c@{\hskip 0.4in}c}
{\includegraphics[scale=0.25]{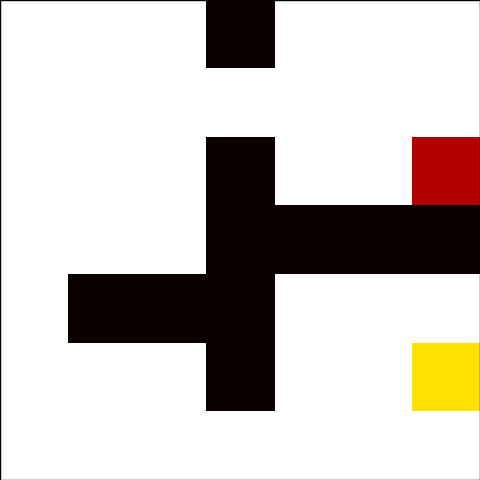}}&
{\includegraphics[scale=0.25]{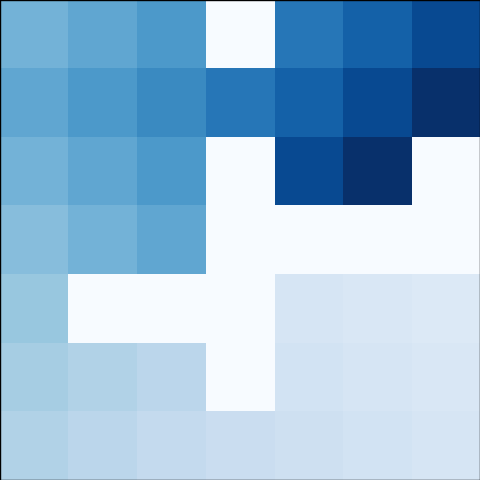}}
\end{tabular}
\caption{\small \emph{Left}: Target task in the grid world with goal position  (\emph{red}), feasible positions (\emph{white}), blocked positions (\emph{black}) and the start position (\emph{yellow}). \textit{Right:} State-Value table learned after convergence for the grid world task. Darker shades represent numerically higher values.}
\label{fig:grid}
\end{figure}

In this domain (see figure \ref{fig:grid} \textit{left}), a 2-D grid world environment is provided to an agent whose objective is to reach a specified destination state starting from a pre-specified starting state. This is an episodic task where the episode ends when the agent reaches to the goal state. The agent receives a positive reward on reaching the destination state and no reward otherwise. If the agent is unable to reach the goal state within $500$ steps, the episodes ends with zero reward.
\begin{figure}[h]
\centering
\begin{tabular}{cccc}
{\includegraphics[scale=0.15]{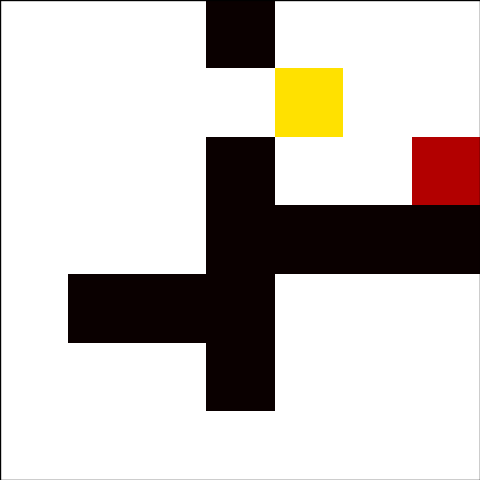}}&
{\includegraphics[scale=0.15]{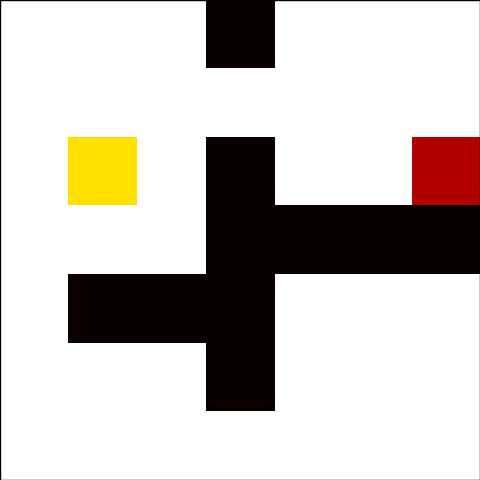}}&
{\includegraphics[scale=0.15]{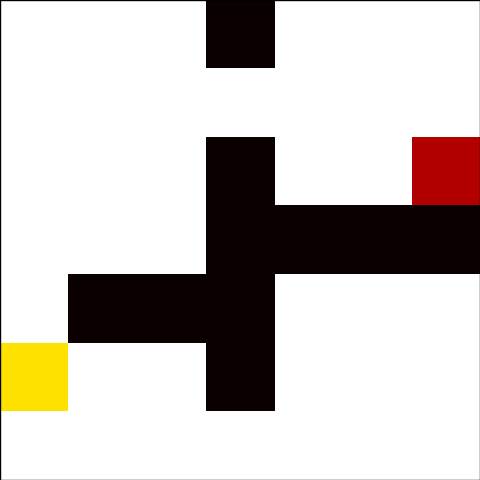}}&
{\includegraphics[scale=0.15]{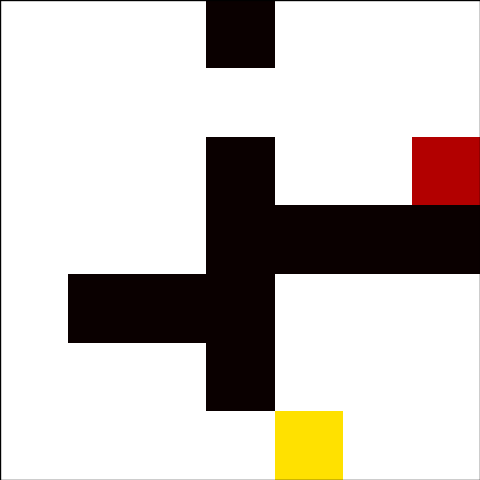}}
\end{tabular}
\caption{\small Training tasks for the grid world task in Figure \ref{fig:grid}.}
\label{fig:grid-sources}
\end{figure}

In the experiment, we used the 2-D grid world illustrated in Figure \ref{fig:grid} \textit{(left)}, which illustrates the target task to be learned by the agent. We created a set of source tasks $\mathcal{T}$ of size 4 as shown in Figure \ref{fig:grid-sources}. The source tasks in this domain are created by changing the starting position of the agent while keeping the goal position and the state space invariant. The learning and transfer strategies of the agent for this domain is similar to that of the maze domain. The agent employs the $Q$-learning algorithm for learning and reies on $Q$-function transfer. We set the discount factor ($\gamma$) to be 0.9 and learning rate to be $0.6$ in the $Q$-learning algorithm.

For evaluation, the target task is learned till convergence. Again, we say that the agent has learned the task if the $Q$-values do not change for 5 consecutive episodes. In \textit{Reward Maximizing Greedy Selector}, the number of steps for evaluating a task to be chosen as the next, is set to $500$; and the training tasks in curriculum are learned till convergence. In \textit{Local Transfer Maximizing Selector}, the number of steps for calculating transfer measure between any two tasks is set to $100$, and the training tasks in the curriculum are learned till convergence as well.

\begin{figure}[h]
\centering
\includegraphics[scale=0.45]{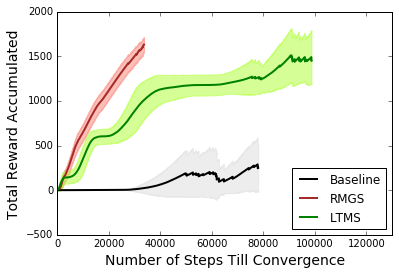}
\caption{\small Reward accumulated versus the number of steps required for convergence for each of the proposed methods (abbreviated in the legend) and the baseline on the grid world task.}
\label{fig:grid-result}
\end{figure}

\begin{figure}[h]
\centering
{\includegraphics[scale=0.45]{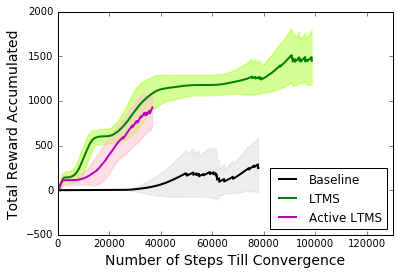}}
\caption{Comparison of the Active LTMS method with the LTMS method and baseline in Figure \ref{fig:maze-result} on the grid world task.}
\label{fig:grid-active}
\end{figure}

Figure \ref{fig:grid-result} shows the results of our active simulator algorithms compared with the baseline. For this domain, we evaluate our methods with the baseline based on the \textit{time to threshold} (see Section \ref{PROBLEM}) criterion, where the threshold performance is defined as the time neede for an agent to converge on the target task. We plot the total reward accumulated by our methods as well as the baseline until the \emph{Q}-function converges on the target task (with Monte Carlo averaging over 30 runs). The state-value function learned using the curriculum is shown in Figure~\ref{fig:grid} \textit{(right)}. The plot in Figure~\ref{fig:grid-result} suggests that the \textit{Reward Maximizing Greedy Selector} converges in lesser number of steps than the baseline, while the LTM selector again takes more number of steps than baseline. Similar to the previous domain, we use an active learning technique as described in Section \ref{PRELIM} to compute inter-task transferability matrix using only a small number of task pairs. In this domain, the inter-task feature $f^{ij} = [1, f^{ij}_1]$ is defined as $f^{ij}_1 = f^i - f^j$ where $f^i$ is the minimum distance to the destination state from the goal state of task $\mathcal{M}^i$. Figure~\ref{fig:grid-active} shows that the number of steps needed reduce significantly using the \textit{Active Local Transfer Maximizing Selector} variant, while the performance and the curricula selected remain the same.

\begin{figure}[h]
\centering
{\includegraphics[scale=0.4]{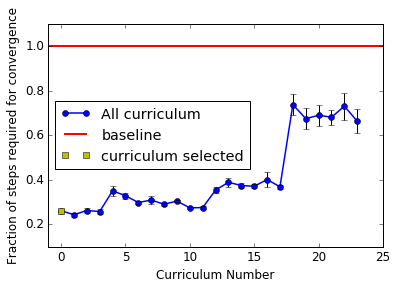}}
\caption{\small Number of steps required for convergence per curriculum (indexed from $0$ to $23$), compared to the baseline(\emph{red}) for the grid world task. The curricula represent all permutations of tasks in Figure~\ref{fig:grid-sources}, ordered lexicographically. \emph{Yellow} squares represent the most frequent (mode) curricula selected by our methods.}
\label{fig:grid-curr}
\end{figure}

Finally, we qualitatively analyze the curriculum proposed by the our algorithms. We compare all possible full-length curricula based on the four training tasks (totaling $24$).  Figure~\ref{fig:grid-curr} shows the convergence time of each curriculum as compared to the baseline and the most frequent curriculum selected by our methods. The ordering corresponding to the most frequent curriculum is shown in order (left to right) in Figure~\ref{fig:grid-sources}.

\subsection{Cart-pole Environment}
\begin{figure}[h]
\centering
{\includegraphics[scale=0.25]{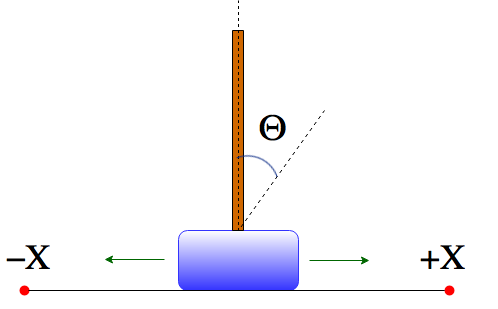}}
\caption{\small The cart-pole environment. An episode ends if either the cart moves out of the horizontal axis bound $[-X,+X]$ or the pole leans outside of the angular bound $[-\Theta, +\Theta]$.}
\label{fig:cart}
\end{figure}

The cart-pole domain is a classical control problem in RL~\cite{barto1983neuronlike}. In this domain, an agent learns to balance a pole on a cart block (see Figure \ref{fig:cart}). The agent's state space is continuous and we denote it by $(\theta, x, \omega, v)$, representing the angular position of pole, position of cart and their time derivatives respectively. These are typically bound by a box in 4 dimensions. The agent can take action on whether to move \textit{left} or \textit{right}, giving us the action space $\{Left, Right\}$. The task is episodic and ends when either the angle $\theta$ goes out of a specified angle bound $[-\Theta, \Theta]$ or if the cart moves out of a specified bound $[-X, +X]$. The agent receives \textit{+1} reward at every step it takes till the episode ends. In other words, the reward of an episode is equal to the number of steps the agent is able to hold the pole on the cart within bounds. The episodes deterministically end after maximum $500$ steps, hence the maximum reward in an episode can be $500$.

In the experiment, we define the target task of the cart-pole task as shown in Figure \ref{fig:cart} with $X=2.4$ and $\Theta=30^{\circ}$ degrees. We create a set of source tasks $\mathcal{T}$ of size 4 are less complex than the target task by varying the bounding box parameters $X$ and $\Theta$. The four source tasks created are $\mathcal{T} = \{(X=4.0, \Theta=60^{\circ}), (X=4.0, \Theta=45^{\circ}), (X=3.2, \Theta=45^{\circ}), (X=3.2, \Theta=30^{\circ})\}$. We modify the \textit{CartPole-v0} environment available from the \textit{OpenAI Gym} platform~\cite{1606.01540} for our experiments. We use the DQN algorithm~\cite{mnih2015human} to train the agent with a fully-connected network consisting of three hidden layers of size $16$ and the \textit{Rectified Linear Unit (ReLU)} non-linearity~\cite{nair2010rectified}. The input and output layers have the sizes equal to the agent's state space and action space respectively. The architecture of the fully-connected network used is $4 \rightarrow 16 \rightarrow 16 \rightarrow 16 \rightarrow 2$. We used the \textit{keras-rl} library \cite{plappert2016kerasrl} to build our algorithms on top of the DQN agent. The experience replay memory size is set to $50000$, the mini-batch size is set to $32$, and the learning rate of the optimization is set to $10^{-4}$. We used the initialization of the the network parameters as the transfer method across tasks for the agent~\cite{parisotto2015actor}.

In this domain, we specify a limit of episodes available for learning to the agent. The baseline learns the target task for all episodes in the budget and our algorithms use the budget to find and then learn on the curriculum obtained as well. For the \textit{Reward Maximizing Greedy Selector}, the number of steps for evaluating a task to be chosen next is set to $200$, and the training tasks in the curriculum are learned for $2000$ steps. For the \textit{Local Transfer Maximizing Selector}, the number of steps for calculating transfer measure between any two tasks is set to $500$ and the training tasks in the curriculum are learned for $2000$ steps.

\begin{figure}[h]
\centering
\includegraphics[scale=0.45]{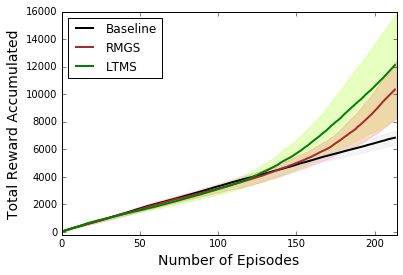}
\caption{\small Reward accumulated versus the number of episodes for each of the proposed methods (abbreviated in the legend) as well as the baseline on the cart-pole task.}
\label{fig:cart-result}
\end{figure}

Figure~\ref{fig:cart-result} shows the results of our active simulator algorithms compared to the baseline. For this domain, we evaluate our methods with the baseline using the \textit{Total Reward} (see Section \ref{PROBLEM}) evaluation criterion where the total reward is the reward accumulated by an agent in some specific number of episodes. We plot the total reward accumulated by our methods as well as the baseline for a fixed number of episodes equal to $\sim 220$ (with Monte Carlo averaging over 30 runs). It can be seen in Figure~\ref{fig:cart-result} that both the \textit{Reward Maximizing Greedy Selector} and the \textit{Local Transfer Maximizing Selector} have accumulated much higher rewards than the baseline in the given number of episodes. Similar results also hold for the domain aware setting (omitted here due to space constraints).

\section{Conclusion}
\label{CONCLUSION}

In this paper, we have defined the online curriculum learning problem as a way to speed up the learning of reinforcement learning agents on a target task. We designed algorithms for two broad settings. 
In the first, no task information is available, and in the second, task dependent features are available. 
Our algorithms were able to choose curricula and train reinforcement learning agents more efficiently compared to training the agents directly on the target tasks for three different domains.  Statistical and computational guarantees for the proposed algorithms are left for future work. Automating the design and choice of source tasks in addition to online curriculum learning is also an interesting direction to pursue.


\bibliographystyle{plain}  
\bibliography{ref}  

\end{document}